# Zero-Shot Multi-Criteria Visual Quality Inspection for Semi-Controlled Industrial Environments via Real-Time 3D Digital Twin Simulation

Jose Moises Araya-Martinez[a,c,*], Gautham Mohan[a,†], Kenichi Hayakawa Bolaños[b,†], Roberto Mendieta[c], Sarvenaz Sardari[a], Jens Lambrecht[c,d], and Jörg Krüger[e]

[a]Mercedes-Benz AG, Future Automotive Manufacturing, Benz-Str. Bau40, 71063 Sindelfingen, Germany
[b]Costa Rica Institute of Technology, School of Computer Engineering, 30050 Cartago, Costa Rica
[c]Gestalt Automation GmbH, Schlesische Straße 26, 10997 Berlin, Germany
[d]Technical University Berlin, Industrial Automation Technology, Pascalstraße 8-9, 10587 Berlin, Germany
[e]Technical University Berlin, Industry Grade Networks and Clouds, Ernst-Reuter-Platz 7, 10587 Berlin, Germany

* Corresponding author. E-mail address: jose_moises.araya_martinez@mercedes-benz.com
† Equal contributions

**Abstract**

Early-stage visual quality inspection is vital for achieving Zero-Defect Manufacturing and minimizing production waste in modern industrial environments. However, the complexity of robust visual inspection systems and their extensive data requirements hinder widespread adoption in semi-controlled industrial settings. In this context, we propose a pose-agnostic, zero-shot quality inspection framework that compares real scenes against real-time Digital Twins (DT) in the RGB-D space. Our approach enables efficient real-time DT rendering by semantically describing industrial scenes through object detection and pose estimation of known Computer-Aided Design models. We benchmark tools for real-time, multimodal RGB-D DT creation while tracking consumption of computational resources. Additionally, we provide an extensible and hierarchical annotation strategy for multi-criteria defect detection, unifying pose labelling with logical and structural defect annotations. Based on an automotive use case featuring the quality inspection of an axial flux motor, we demonstrate the effectiveness of our framework. Our results demonstrate detection performace, achieving intersection-over-union (IoU) scores of up to 63.3% compared to ground-truth masks, even if using simple distance measurements under semi-controlled industrial conditions. Our findings lay the groundwork for future research on generalizable, low-data defect detection methods in dynamic manufacturing settings.

## 1. Introduction

The importance of quality inspection for early error detection and cost-effective production has long been recognized in foundational manufacturing strategies such as Zero-Defect Manufacturing [1] and Lean Production [2]. However, despite ongoing research in Deep Learning (DL) for manufacturing [3], several challenges continue to hinder the widespread adoption of automated quality inspection [4]. In particular, the need for extensive data collection and annotation to train modern DL models on proprietary datasets significantly slows their deployment [5]. Moreover, limited expertise in modeling technologies —essential for effective process simulation through DTs— especially affects small and medium-sized enterprises [6, 7]. Additionally, the absence of a unified format for representing common defect types (e.g., positional, structural, and logical) hampers the modular deployment of multi-criteria quality inspection systems [6, 7].

To address these limitations and promote the democratization of industrial quality inspection, this work introduces the following key contributions:

- We introduce a novel zero-shot framework for industrial multi-criteria quality inspection in semi-controlled environments. Our method enables real-time comparison with DTs generated from Computer-Aided Design (CAD) data, eliminating the need for manual defect annotation.
- Given the real-time constraints of early error detection, we benchmark three widely used rendering engines and propose runtime optimizations for fast DT generation enriched with quality-relevant, multi-criteria annotations.
- To address the challenge of noisy object poses in semi-controlled settings, we incorporate a refinement stage that improves detection accuracy even under uncertain or imperfect input conditions.
- Finally, we propose a hierarchical and extensible defect annotation format that unifies positional, structural, and logical inspection tasks. Our format builds upon established standards such as BOP and COCO, advancing the standardization of multi-criteria defect data representation.

**2. Related Works**

This section summarizes major previous research outcomes that motivated our work. We mention approaches for real-time DT creation building upon the concept of Cyber Physical Systems. Furthermore, we mention previous attempts to standardize multi-criteria annotations in a unified format, as well as common metrics for multimodal, color- and geometry-based image quality.

*2.1. Structural and Logical Inspection via Cyber-Physical Systems and DTs*

Advances in simulation-to-reality (sim-to-real) DL training, spark initiatives to automate offline large-scale data generation and annotation following Domain Randomization or Domain Adaptation approaches [8, 9, 10, 11]. Furthermore, the Cyber-Physical System concept, employed in the generation of DTs with quality-relevant characteristics [7], enables the generation of real-time ground truths to visual-based multi-criteria inspection tasks.

DTs can provide high fidelity virtual references [12], these virtual references can be employed to compare the manufactured state of an object with its expected state. They can also incorporate varying levels of semantic information, enabling flexible adaptation to different manufacturing processes including inspection.

In this context, Feng et al. [13] propose an occlusion-aware, visible-only DT framework. Occlusion awareness in their system is achieved using a Z-buffer, a standard feature supported by most rendering engines [14]. Their work highlights the effectiveness of part focused inspection compared to a global inspection, and by constraining the inspection to visible surfaces, the method avoids false positives from self-occlusion.

*2.2. Real-Time Rendering Engines for DT Creation*

Rendering in the order of milliseconds is essential to DT systems, enabling responsive visualization and interaction with virtual counterparts of physical assets. Several open-source frameworks have emerged as viable options for research.

Open3D supports real-time rendering of point clouds, meshes, and RGB-D data with GPU acceleration. Widely used in robotics and computer vision, it integrates well with SLAM systems, object reconstruction pipelines, and DL workflows, making it suitable for DT applications [15].

Trimesh is a lightweight Python library for mesh manipulation and visualization, known for fast mesh loading, transformation utilities, and tool interoperability [16]. When paired with Pyrender, which provides an OpenGL-based rendering backend, it enables real-time scene composition and pose-aware rendering [17].

BlenderProc leverages Blender for procedural scene generation and photorealistic rendering [18], commonly used for synthetic data generation and 6D pose estimation in benchmarks like BOP [21]. Though not a real-time engine, its realism and control make it useful for simulation and training.

While none of these tools are a complete solution on their own, each offers distinct advantages depending on the specific requirements of the DT pipeline, such as rendering speed, visual realism, or pose integration.

*2.3. Annotation of Multi-criteria Quality Inspection Datasets*

Bergmann et al. [19] classify anomalies as either structural or logical. Structural anomalies are defined as features not occurring in anomaly-free data, while logical anomalies are defined as situations not-conformant with logical constraints in the data. To cover this anomaly classification, MVTec Logical Constraints Anomaly Detection (MVTec LOCO AD) dataset is proposed with annotated pixel-wise structural and logical anomalies across five categories in an industrial scenario.

The MVTec LOCO AD is further expanded into the LOCO-Annotations [20], which includes information about the causes of logical anomalies. It categorizes the logical anomalies into quantity anomalies, size anomalies, position anomalies, and matching type anomalies, thus giving more detail on the nature of the anomaly. The added information is proposed to be used for further model tuning [20].

These datasets have the limitation of being only focused on 2D information and working under the assumption that the objects to be analyzed will always be in a fixed position aligned to the camera [19]. For analysis of a pose-agnostic defect detection model we recognize the need for RGB and depth data from multiple points of view.

## 2.4. Evaluation of Pose Refinement Quality

Evaluating 6D pose estimation accuracy, particularly after refinement with algorithms like Iterative Closest Point (ICP), requires standardized metrics. Two common approaches are distance-based pose error metrics and success rates across multiple accuracy thresholds [21].

A widely used metric is the Average Distance of Model Points (ADD), which computes the mean Euclidean distance between model points transformed by the estimated pose and those transformed by the ground truth pose [22]. The result represents the average deviation of the object's transformed geometry from its true pose. Lower ADD values indicate better alignment and are effective for assessing refinement quality.

Rather than relying on a single threshold, accuracy is evaluated across multiple thresholds each representing a different tolerance for pose error.

## 3. Methodology and Implementation

In this section, we summarize our approach to demonstrate our key findings. In Subsection 3.1 we outline our proposed framework and basic quality metrics for zero-shot, multi-criteria quality inspection. Furthermore, in Subsection 3.2, our hierarchical annotation format for labeling of logical and structural defects is exposed. Finally, in Section 3.3 we introduce the method used for refinement of imperfect poses.

### 3.1. Zero-Shot Quality Inspection Framework

Fig. 1 shows an overview of our proposed framework for zero-shot general quality inspection with detection capabilities on multiple logical and structural defects. In the offline phase, the generation of synthetic data to train an object detection model closely follows our previous work on Domain Randomization and Domain Adaptation for industrial object detection [10, 11]. In the online phase, this trained model is used along with a pose estimation model to create a real-time DT, which can be used as a reference for defect detection.

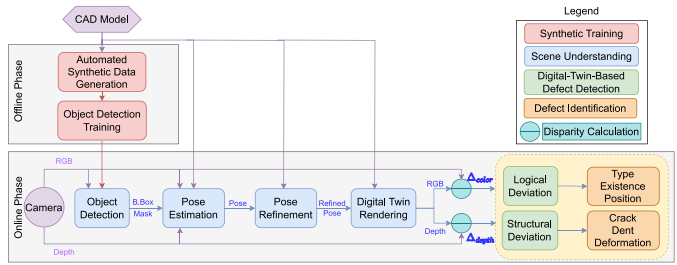

Fig. 1. Framework for zero-shot quality inspection with offline training and online inference. In the offline stage, synthetic data is used to train YOLO. During the online operation, scene understanding mitigates environmental variability to enable real-time CAD-based digital twin rendering, allowing defect detection without manual annotations.

The different stages in the online phase are discussed below–

1. Object detection – A pretrained YOLO object detector [23] is used for detecting the object in the scene. The bounding box is converted to a rectangular mask.
2. Pose estimation – A FoundationPose model [24] is initialized with a CAD model and a detection mask of the object. The RGB and depth image from the camera is input to the model. The pose can be further refined with methods like ICP.
3. DT Rendering - A DT is rendered using Trimesh and Pyrender libraries. This returns the RGB and Depth image of the object in the current pose.
4. Depth comparison - The rendered depth, $D_{render} \in \mathbb{R}^{m \times n}$, is compared against the actual depth, $D_{real} \in \mathbb{R}^{m \times n}$, per pixel to generate the geometric deviation, $\Delta_{depth} \in \mathbb{R}^{m \times n}$.

$$\Delta_{depth}[i,j] = \{\| D_{render}[i,j] - D_{real}[i,j] \|_1 : 0 \leq i < m, 0 \leq j < n\} \quad (1)$$

5. Color comparison - The RGB image, $C_{render} \in \mathbb{R}^{m \times n \times 3}$, from the renderer and the camera, $C_{real} \in \mathbb{R}^{m \times n \times 3}$, is converted to LAB format and the pixel-wise channel deviation given by $\Delta_{color} \in \mathbb{R}^{m \times n}$ is calculated as Euclidean distance between every pixel-pair.

$$\Delta_{color}[i,j] = \{\| C_{render}[i,j] - C_{real}[i,j] \|_2 : 0 \leq i < m, 0 \leq j < n\} \quad (2)$$

The depth comparison accounts for self-occlusions as depth is generated from the Z-buffer by the rendering engine.

### 3.2. Method for Generation and Annotation of Multi-Criteria, Hierarchical Ground Truth Data

To validate our annotation-free multi-criteria quality inspection framework, manual annotation of ground truth defects is required. Thus, we implemented a data creation setup consisting of an Ensenso C-57M camera mounted on top of an UR-5 robot arm. The camera provides a depth accuracy of 0.2 mm and a resolution of 2 MP. An anomaly is manually produced, then the robot arm moves to a predefined set of waypoints and triggers the camera in each of them. With each camera trigger, aligned depth and color images are obtained. The test object consists of a 3D printed Axial flux motor containing a base plate and 12 coils. Coils are mounted on slots on top of the plate, and they have a unique order.

The resulting image pair is first annotated with the object pose of the whole assembly, following the BOP format [25]. Then the color images are annotated for anomalies occurring in the coils. The annotated dataset results in the structure depicted in Fig. 2(a). Additionally, Fig. 2(b) shows the main two defect super-categories foreseen in our annotation format, i.e. logical and structural, as well as their sub-categories. Furthermore, the outermost level in the sunburst diagram of Fig. 2(b) offers examples of possible defects under each sub-category. Logical anomalies are labelled as polygons while structural are labelled as pixel masks.

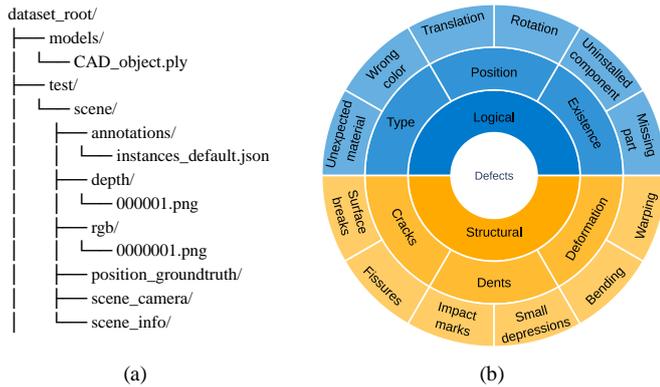

Fig. 2. The proposed format for multi-criteria defect annotations. In (a), the folder and file structure are depicted. In (b), the super categories for "Structural" and "Logical" defects, as well as their child's nodes are shown.

As shown in Fig. 3, deformation involves an increase in material and is simulated with additional 3D printed volume of either 4 cm$^3$ or 1 cm$^3$. Furthermore, crack anomalies are fractures, either 3 mm or 10 mm deep. Existence identifies a missing coil, annotated at the position it should have. Position anomalies identify an existing, but misaligned coil. Type identifies existing coils, correctly placed on the plate, but placed out of order (in this case 2 coils per image are labelled).

In total, for each waypoint position 85 images are obtained from these 48 are structural anomalies while 37 are logical anomalies (four of the images contain double logical annotations, so this brings the number of logical anomaly annotations to 41).

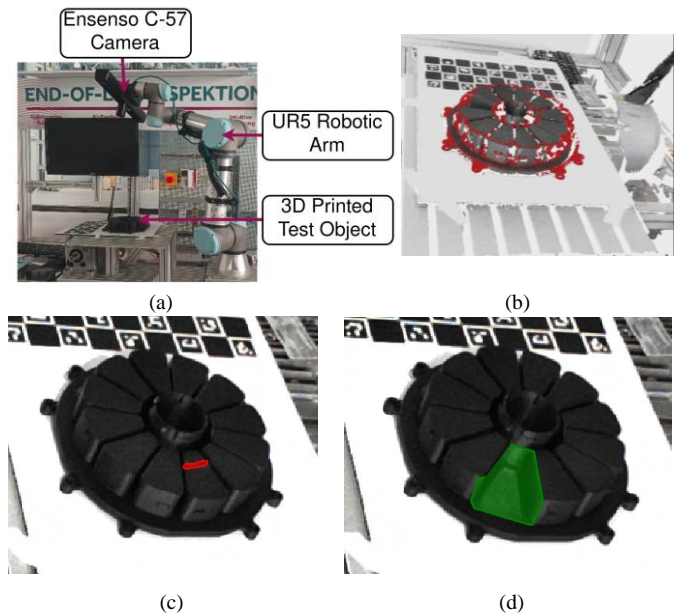

Fig. 3. Robot-based setup for pose and defect annotation. (a) Lab environment with robot, 3D camera, and test object. (b) Contour overlay for pose validation. (c) Sample structural defect (crack) with segmentation mask. (d) Logical defect (existence) shown as polygon mask.

### 3.3. Increasing Detection Robustness via Pose Refinement

Initial pose estimates may contain inaccuracies, impacting the ground truth quality of the derived DT. To address this, we propose using pose refinement using the ICP algorithm to improve the accuracy of estimated poses by aligning them more closely with the known ground truth of the object.

Our pose refinement methodology consists of the following steps:

1. Initial Pose Estimation: We begin with an estimated pose from FoundationPose, which may contain both rotational and translational errors.
2. Point Cloud Generation: We generate a target point cloud from the scene's depth image using the camera's intrinsic parameters. For each pixel, we compute its corresponding 3D coordinates and filter out invalid depth values.
3. Source Model Preparation: The 3D CAD model of the target object is converted into a point cloud and positioned according to the initial pose estimate.
4. Point Cloud Registration: We employ Open3D's implementation of ICP. Point-to-plane ICP is used when surface normal information is available, falling back to point-to-point ICP when necessary.
5. Transformation Refinement: The resulting transformation from ICP is combined with the initial pose to produce the refined pose estimate.

To assess the effectiveness of our pose refinement approach, we employ several complementary metrics. Rotational error is measured in degrees using the Frobenius norm between rotation matrices, while translational error is calculated as the Euclidean distance between translation vectors. The ADD captures the mean deviation between model points transformed by the estimated and ground truth poses, with lower values indicating better alignment. Additionally, we report success rates across predefined distance thresholds, which reflect how closely the refined poses match the ground truth.

### 4. Results

The proposed zero-shot quality inspection framework was evaluated across three key aspects - real-time DT rendering performance, defect detection, and pose refinement. The results of the experiments conducted are summarized in the sections below.

#### 4.1. Benchmark of Real-Time Rendering Engines

The different rendering tools were tested on a machine with Intel® Core™ i7-8750H CPU running at 2.20 GHz and a GeForce RTX 2080 Mobile GPU. The run-time comparison from the experiments is highlighted in Fig. 4. In our experiments, Open3D and the combination of Trimesh + Pyrender demonstrated comparable performance, with average rendering times of approximately 0.01 s and writing times around 0.03 s per frame.

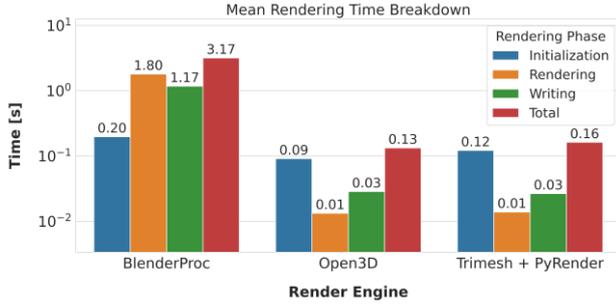

Fig. 4. Runtime benchmark of BlenderProc, Open3D and Trimesh + PyRender as real-time DTs rendering engines. Values represent the mean of 100 runs while rendering Object 1 of LINEMOD [26].

As the initialization overhead occurs only once and is negligible in the context of repeated rendering tasks, runtime performance was not a decisive factor in tool selection. Instead, the choice to adopt Trimesh and Pyrender was driven by practical and functional considerations. This combination offered a more streamlined interface for applying 6D object poses, rendering scenes using intrinsic camera parameters, and managing object transformations, all of which aligned more directly with the pipeline's requirements. Additionally, its usability and flexibility in handling pose-based rendering made it better suited for rapid prototyping and integration in the context of DT creation and evaluation. Furthermore, while rendering a LINEMOD object, Trimesh and Pyrender exhibit a total RAM consumption 25% lower than Open3D, making it possible to allocate real-time rendering in systems with as low as 320MB available RAM.

### 4.2. Pose Refinement

For assessing the effectiveness of the ICP-based pose refinement, total of 1000 synthetically perturbed object poses were created. Each unique pose is corrupted with a combination of rotation and translation errors within a range of 0°-10° and 0-10 mm, correspondingly. The resulting homogeneous noise distribution aims at simulating typical pose deviations. The bar graph in Fig. 5 displays the distribution of pose errors in terms of ADD after correction using ICP. The green bars represent ICP-corrected pose counts within each accuracy bracket.

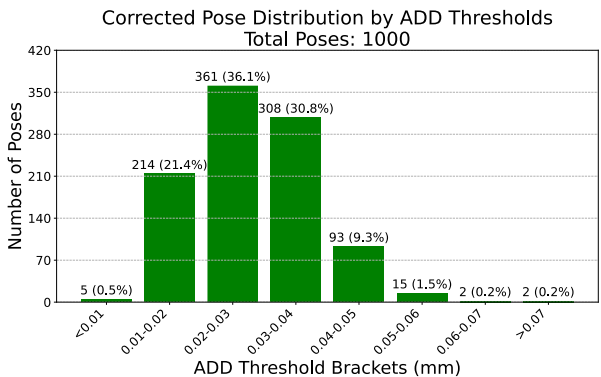

Fig. 5. Pose refinement tests, summarized as the distribution of pose estimation accuracy following ICP refinement, segmented by precision thresholds.

A pose is considered successful at a given threshold when its ADD is below that threshold value. The data demonstrates that when starting with initial pose estimates within moderate error bounds (maximum 10° rotation and 10 mm translation error), ICP consistently refines these poses to achieve sub-millimeter accuracy. Most notably, 100% of poses are successfully corrected below the 1 mm ADD threshold, which demonstrates a level of precision well-suited for high-accuracy industrial inspection tasks.

Several considerations should be noted. ICP is a local optimization algorithm, meaning its effectiveness diminishes with larger initial errors outside the convergence basin. Scene complexity should be considered, as the results were obtained in controlled conditions, and real-world industrial scenes with occlusions, clutter, or sensor noise may lead to less optimal outcomes. Finally, object-specific performance is also important, as results can vary for objects with different geometric characteristics.

These findings suggest that when applied to well-structured problems with appropriate initial conditions, ICP refinement can dramatically improve pose estimation accuracy, potentially enabling applications requiring high geometric precision.

### 4.3. Quality Inspection in Color and Depth Space

The existence and color category from the annotations proposed in Section 3.2 were taken to prove the effectiveness of the proposed framework quantitatively. Fig. 6 illustrates the qualitative results for existence, color and deformation defects. Here, the existence check is performed using depth disparity images with Equation 1 and the color check is performed using color disparity images, as per Equation 2. The results closely match the ground truth mask.

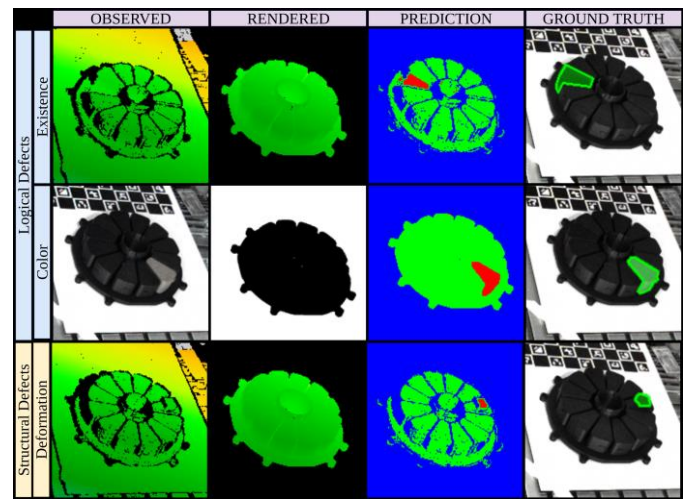

Fig. 6. Qualitative results on zero-shot color (logical), existence (logical), and deformation (structural) defects. The real-time DT, rendered as RGB-D, serves as reference for volume and color deviations.

Table 1 provides the IoU values for existence and color defects. The relatively low values here can be attributed to the noise in the depth values and the change in color intensity because of effects of light and texture properties of the object.

Table 1. Results of Depth and Color Deviation in terms of mean IoU with Annotation mask.

| Anomaly Type | Mean IoU |
|---|---|
| Existence | 63.3% |
| Color | 62.9% |

## 5. Conclusions and Future Work

In this paper we introduce a novel framework for multi-criteria defect detection. We benchmarked the performance of three commonly used rendering tools on speed and memory, ultimately choosing Trimesh + Pyrender for our pipeline. The results also highlight the importance of pose refinement for reducing detection errors due to pose mismatches. The rendering method was combined with pose estimation to generate a real-time DT as RGB-D reference for defect detection. We also introduced a hierarchical extensible annotation format for multi-criteria defect detection which we used to evaluate the efficacy of the proposed pipeline.

The current implementation utilizes DL methods sparsely for overcoming the varying nature of semi-controlled environments in the scene-understanding stage. However, the results advocate for the broader integration of DL methods as future work, particularly to compute color disparities between DTs and real objects. We also recognize the need to perform defect classification into the classes specified in the annotation; this would provide the user with additional information to take informed action towards error correction.

## Acknowledgements

This work has been funded by the German Federal Ministry for Economic Affairs and Climate Action based on a resolution of the German Bundestag, financed by the European Union.